\documentclass{article}
\usepackage[preprint]{spconf,amsmath,graphicx}

\usepackage{times}
\usepackage{epsfig}
\usepackage{graphicx}
\usepackage{amsmath}
\usepackage{amssymb}

\usepackage{framed,multirow}

\usepackage{amssymb}
\usepackage{latexsym}

\usepackage{amsmath}
\usepackage{makecell}
\usepackage{multirow}
\usepackage{booktabs}

\title{Cross-Age Contrastive Learning for Age-Invariant Face Recognition}
\newcommand\Mark[1]{\textsuperscript#1}
\name{Haoyi Wang\Mark{1}\thanks{Copyright 2024 IEEE. Published in ICASSP 2024 - 2024 IEEE International Conference on Acoustics, Speech and Signal Processing (ICASSP), scheduled for 14-19 April 2024 in Seoul, Korea. Personal use of this material is permitted. However, permission to reprint/republish this material for advertising or promotional purposes or for creating new collective works for resale or redistribution to servers or lists, or to reuse any copyrighted component of this work in other works, must be obtained from the IEEE. Contact: Manager, Copyrights and Permissions / IEEE Service Center / 445 Hoes Lane / P.O. Box 1331 / Piscataway, NJ 08855-1331, USA. Telephone: + Intl. 908-562-3966.}\qquad Victor Sanchez\Mark{2} \qquad Chang-Tsun Li\Mark{3}}
\address{\Mark{1}School of Engineering, Computing and Mathematics, University of Plymouth, Plymouth, UK \\ 
         \Mark{2}Department of Computer Science, The University of Warwick, Coventry, UK \\ 
         \Mark{3}School of Information Technology, Deakin University, Waurn Ponds, Australia \\
         \texttt{haoyi.wang@plymouth.ac.uk}}
\begin{document}
%
\maketitle
\begin{abstract}
Cross-age facial images are typically challenging and expensive to collect, making noise-free age-oriented datasets relatively small compared to widely-used large-scale facial datasets. Additionally, in real scenarios, images of the same subject at different ages are usually hard or even impossible to obtain. Both of these factors lead to a lack of supervised data, which limits the versatility of supervised methods for age-invariant face recognition, a critical task in applications such as security and biometrics. To address this issue, we propose a novel semi-supervised learning approach named \textbf{C}ross-\textbf{A}ge \textbf{Con}trastive Learning (CACon). Thanks to the identity-preserving power of recent face synthesis models, CACon introduces a new contrastive learning method that leverages an additional synthesized sample from the input image. We also propose a new loss function in association with CACon to perform contrastive learning on a triplet of samples. We demonstrate that our method not only achieves state-of-the-art performance in homogeneous-dataset experiments on several age-invariant face recognition benchmarks but also outperforms other methods by a large margin in cross-dataset experiments.
\end{abstract}
%
\section{Introduction}
\label{sec:intro}
Age-invariant face recognition (AIFR) aims to recognize the identity of subjects regardless of their age. Different from conventional face recognition, AIFR needs to consider the intra-class variance caused by age information. A robust solution for AIFR can be used in various biometrics and forensics applications, such as tracking a person-of-interest, including missing children, people with dementia, or suspects over a span of several years \cite{wang2020using}.

Most existing AIFR methods attempt to solve the problem under supervised settings. For example, \cite{shakeel2019deep, wen2016latent} aim to learn and extract age-invariant features directly from input images, while generative models \cite{zhao2020disentangled, zhao2019look} synthesize samples that match the target age before feature extraction. However, cross-age facial images of the same subject are extremely difficult to collect. As a result, existing noise-free age-oriented face datasets are small in size, with limited samples per subject \cite{ricanek2006morph}, which could degrade the performance of supervised approaches. To address this, \cite{xu2017age} tries to tackle the problem in an unsupervised manner by implementing an customized auto-encoder. However, this method requires image pairs of the same subject as the input and may not be efficient when the number of images per subject is limited. Furthermore, in real scenarios, images of the same subject at different ages are often hard or even impossible to obtain, limiting the versatility of this and other supervised approaches.

\begin{figure}[t]
\begin{center}
\includegraphics[width=0.95\linewidth]{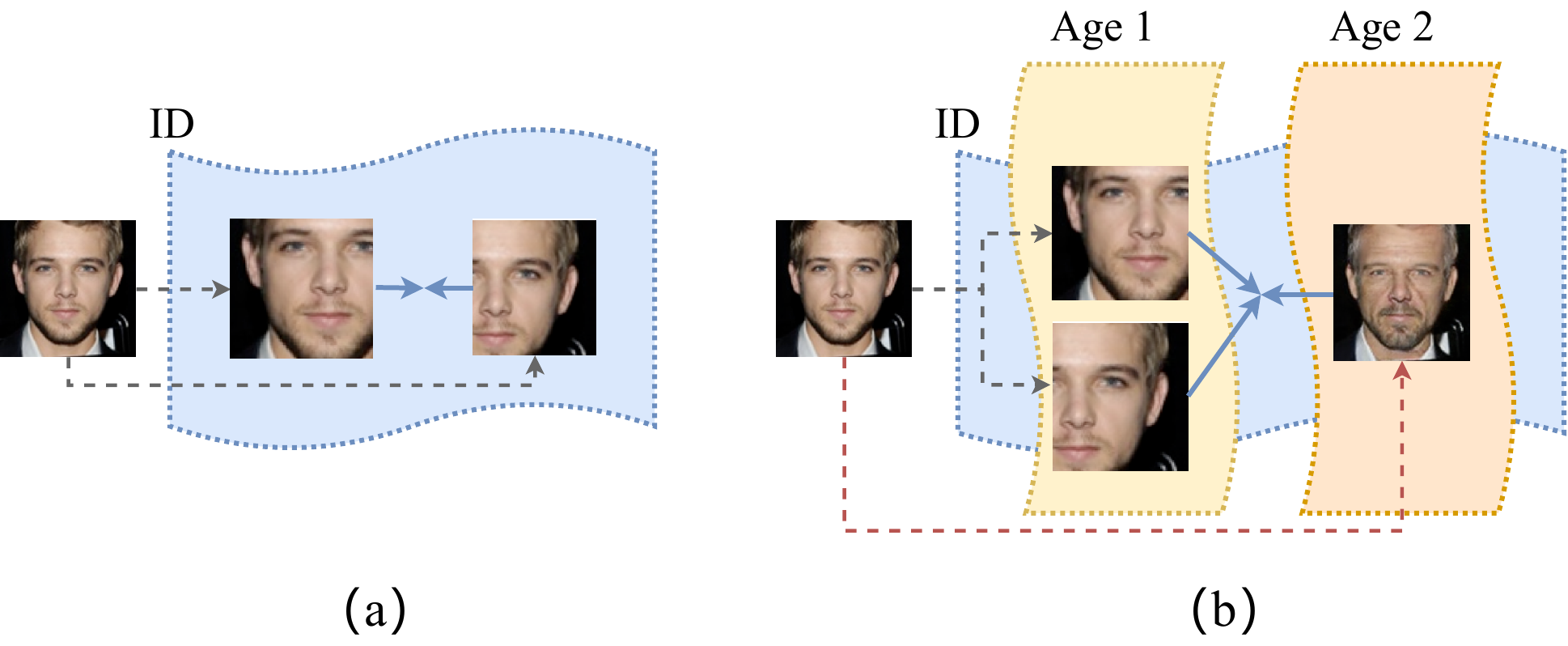}
\end{center}
   \caption{Data augmentation strategy used in (a) conventional contrastive learning, where two augmented samples are used to learn the shared features representing the identity within the input image and (b) CACon, where an additional sample is generated by a face synthesis model and used to learn the common identity features across different ages.}
\label{fig:demo}
\end{figure}

In this paper, we propose a method called \textbf{C}ross-\textbf{A}ge \textbf{Con}trastive Learning (CACon) to tackle the AIFR problem in a semi-supervised way. Specifically, we adopt the idea of contrastive learning to maximize the similarity between features extracted from a pair of augmented samples from the same input image. Unlike conventional contrastive learning methods \cite{chen2020simple}, we use an additional face synthesis model to synthesize the third augmented sample within a random age group. By maximizing the similarity among features from samples derived from the same image but within different age groups, common identity features can be learned. Examples of augmented samples in conventional contrastive learning and CACon are depicted in Fig. \ref{fig:demo}. We further modify the conventional contrastive loss to fit this triplet setting.

Our contributions can be summarized as follows:
\begin{itemize}
\item We propose a contrastive learning approach named CACon to tackle the AIFR problem by utilizing cross-age samples. To the best of our knowledge, our work is the first to successfully apply the idea of contrastive learning to the AIFR problem.
\item We propose a new contrastive loss to fit the training strategy with three augmented samples.
\item We evaluate the proposed method on several AIFR benchmark datasets to show that our method can achieve promising performance in both homogeneous-dataset and cross-dataset settings.
\end{itemize}

The rest of this paper is organized as follows. In Section 2, we present details of CACon including the augmentation strategy and the modified contrastive loss. In Section 3, we explain the experimental settings and discuss the performance results on several AIFR benchmark datasets. Finally, we conclude our work in Section 4.

\section{Cross-Age Contrastive Learning}

In this section, we provide a detailed explanation of the proposed CACon by first formulating the contrastive AIFR task. Next, we delve into the data augmentation process involved in our method, followed by a discussion of the modified contrastive loss.

\begin{figure}[t]
\begin{center}
\includegraphics[width=0.95\linewidth]{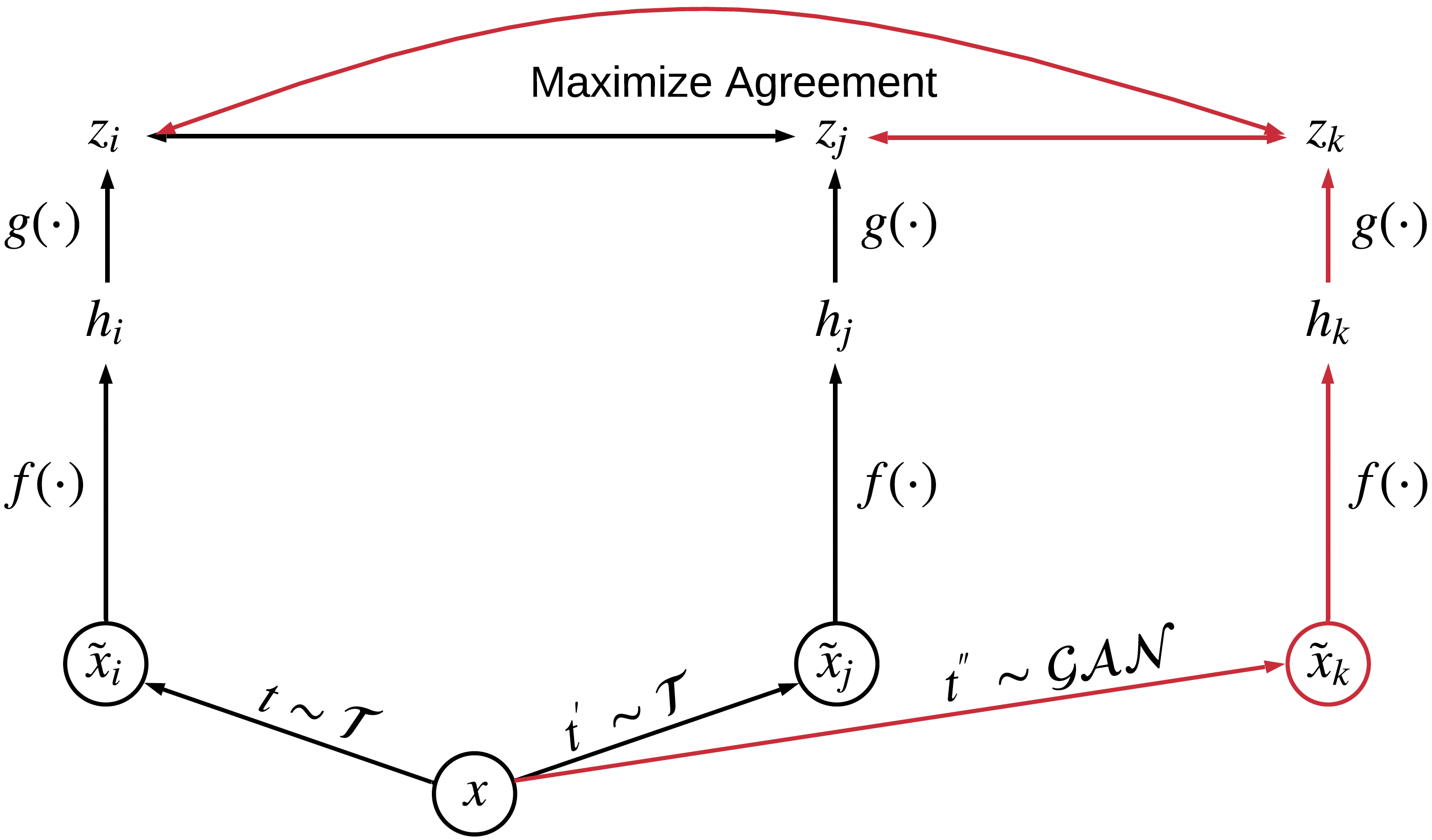}
\end{center}
   \caption{Architecture of the proposed CACon where the black arrows indicate the conventional contrastive learning and the red arrows illustrate the additional path used to learn age-invariant identity features.
   } 
\label{fig:architecture}
\end{figure}  

\subsection{Problem Formulation}

We adhere to the general procedure of contrastive learning as described in \cite{chen2020simple}, which involves two stages. In the first stage, during pre-training, we utilize no labels associated with input images to train an age-invariant identity feature extractor. In other words, given an input image $x$, our goal is to acquire its identity features that remain unaffected by age variations. Subsequently, in the second stage, we fine-tune the final linear layer using labels to enhance task-specific performance. It is worth noting that in this paper, we only demonstrate the possibility of using contrastive learning to solve the AIFR problem. Thus, one of the simplest forms is adopted in our CACon. More advanced methods can be integrated into our approach based on use cases.

As depicted in Fig. \ref{fig:architecture}, in conventional contrastive learning, a pair of augmented samples is generated through a stochastic data augmentation process, $\mathcal{T}$. The two augmented samples are treated as a positive pair and denoted as $\tilde{x}_i$ and $\tilde{x}_j$. A feature extractor $f(\cdot)$ then produces multi-dimensional features $h_i$ and $h_j$ from these augmented samples. Subsequently, $h_i$ and $h_j$ are fed into a projection head $g(\cdot)$, which is used to generate feature vectors $z_i$ and $z_j$. This procedure can be summarized by the following equations:
\begin{equation}\label{general}
    z_i = g(f(\tilde{x}_i)),
\end{equation}
and 
\begin{equation}\label{general}
    z_j = g(f(\tilde{x}_j)),
\end{equation}
where $f(\tilde{x}_i)$ is equivalent to $h_i$ and $f(\tilde{x}_j)$ is equivalent to $h_j$. To extract multi-dimensional features, $f(\cdot)$ is typically formulated as CNNs. To generate one-dimensional feature vectors, $g(\cdot)$ is usually formulated as a stack of linear layers.

This conventional workflow can learn robust identity features for a standard face recognition task, where age variation is not taken into account. To learn age-invariant identity features, the model needs to capture the common identity features across different age groups. To achieve this, in CACon, we employ additional augmented samples $\tilde{x}_k$ that is synthesized using an age-oriented face synthesis model. By maximizing the similarities among features representing the same subject but within different age groups, the model can develop cross-age capabilities and learn age-invariant identity features. The agreement maximization is then performed among a set of three features: $z_i$, $z_j$, and $z_k$.


\subsection{Data Augmentation}

For $x_i$ and $x_j$, we follow the stochastic data augmentation process described in \cite{chen2020simple}. Specifically, this process includes random cropping, resizing to ensure the spatial dimensions of the cropped image match the original, random color distortions, and random Gaussian blur.


For $\tilde{x}_k$, we adopt the age-oriented face synthesis model from \cite{wang2021age}. Instead of using commonly employed L1 or L2 loss to preserve identity information as in other works, \cite{wang2021age} proposed a new identity-preserving loss based on the triplet loss. This new loss function can effectively retain identity features from a multi-task feature extractor capable of extracting disentangled age and identity features. The identity-preserving loss in \cite{wang2021age} can be formulated as:
\begin{equation}\label{adversarial-triplet_modified}
\begin{aligned}
    \mathcal{L}_{AT} (a,p,n) = &\sum_{\substack{t=1}}^T\sum_{\substack{s=1}}^S[m + \max_{\substack{p}}Dist_{a,p} - \min_{\substack{n}}Dist_{a,n}]_+ \\
    &+ [Dist_{n,p} - \min_{\substack{n}}Dist_{a,n}],
\end{aligned}
\end{equation}
where $a$, $p$, and $n$ represent the triplet. $t$ indicates the class index and $s$ is the image index in one batch. 

During data augmentation for pre-training, a label used by the face synthesis model is randomly generated so that CACon can utilize images from different age groups. Moreover, we allow each age group to span 5 years instead of the four coarser groups used in \cite{wang2021age}. This finer age group granularity can help our model capture subtler age variations.

\subsection{Modified Contrastive Loss}

As previously mentioned, given a pair of feature vectors $z_i$ and $z_j$, the contrastive loss aims to maximize the similarity between them. We adopt the commonly used cosine similarity which is calculated as follows:
\begin{equation}\label{sim}
\begin{aligned}
    sim(\tilde{z}_i, \tilde{z}_j) = \frac{\tilde{z}_i^T \cdot \tilde{z}_j}{||\tilde{z}_i||||\tilde{z}_j||},
\end{aligned}
\end{equation}
where $\cdot$ denotes the dot product. $||\tilde{z}_i||$ and $||\tilde{z}_j||$ are L2 normalized feature vectors. In addition to maximizing the similarity between features representing the same subject, we also aim to minimize the similarity between features extracted from different images. To achieve this, the normalized temperature-scaled cross-entropy loss (NT-Xent) has been employed in previous works \cite{chen2020simple} as the contrastive loss. The NT-Xent loss for a pair of features is formulated as:
\begin{equation}\label{nt-xent}
\begin{aligned}
    \mathcal{L}_{NT-Xent}(i,j) = -log\frac{exp(\frac{sim(z_i, z_j)}{\tau})}{\sum_{b=1}^{2B}\mathbf{1}_{[b\ne i]}exp(\frac{sim(z_i, z_b)}{\tau})},
\end{aligned}
\end{equation}
where $\mathbf{1}_{[n\ne i]}$ equals 1 iff $n\ne i$, otherwise 0. $\tau$ represents the temperature parameter \cite{wu2018unsupervised}, and $z_b$ denotes an augmented sample from other images within the same batch. $z_i$ acts as the anchor and $z_j$ is the positive sample. Given a batch size of $B$, there are $2B$ augmented samples in conventional contrastive learning.

In CACon, since a third augmented sample $z_k$ is considered and we want to maximize similarities among the triplet, Eq. \ref{nt-xent} can be modified as:
\begin{equation}\label{nt-xent_modified}
\begin{aligned}
    \mathcal{L}_{NT-Xent}(i,j,k) = -log\frac{exp(\frac{sim(z_i, z_j)}{\tau})+exp(\frac{sim(z_i, z_k)}{\tau})}{\sum_{b=1}^{3B}\mathbf{1}_{[b\ne i]}exp(\frac{sim(z_i, z_b)}{\tau})}.
\end{aligned}
\end{equation}

For computational simplicity, in the above equation, we only maximize the similarity between the anchor and two positive samples. As each sample will be treated as an anchor within one batch, ultimately, the similarities among the triplet can be maximized.




\section{Experiments}

\subsection{Datasets}

We evaluate our method on three commonly used AIFR benchmark datasets: the FG-NET dataset \cite{cootes2008fg}, the MORPH II dataset \cite{ricanek2006morph}, and the CACD-VS dataset \cite{chen2014cross}.

\begin{table}
\begin{center}
\label{fg_homogenous}
\caption{Comparison between state-of-the-art methods and CACon for homogeneous-dataset evaluations on the FG-NET dataset.}
\begin{tabular}{l@{\hskip 0.3in}c@{\hskip 0.3in}c@{\hskip 0.3in}c}\toprule
\hfil{Method} & \hfil{LOIO} & \hfil{MF1} & \hfil{MF2}  \\ \midrule
\hfil{LF-CNN \cite{wen2016latent}} & \hfil{88.10} & \hfil{-} & \hfil{-} \\
\hfil{OE-CNN \cite{wang2018orthogonal}} & \hfil{-} & \hfil{58.21} & \hfil{53.26} \\
\hfil{DM \cite{shakeel2019deep}} & \hfil{92.23} & \hfil{-} & \hfil{-} \\
\hfil{AIM \cite{zhao2019look}} & \hfil{93.20} & \hfil{-} & \hfil{-} \\
\hfil{DAL \cite{wang2019decorrelated}} & \hfil{94.50} & \hfil{57.92} & \hfil{60.01} \\
\hfil{MT-MIM \cite{hou2021disentangled}} & \hfil{94.21} & \hfil{-} & \hfil{-} \\
\hfil{MTLFace \cite{huang2021age}} & \hfil{94.78} & \hfil{57.18} & \hfil{-} \\
\hfil{MFNR-LIAAD \cite{truong2023liaad}} & \hfil{\textbf{95.11}} & \hfil{60.11} & \hfil{-} \\
\midrule
\hfil{SimCLR \cite{chen2020simple}} & \hfil{90.36} & \hfil{54.00} & \hfil{52.52} \\
\hfil{CACon (ours)} & \hfil{{94.61}} & \hfil{\textbf{64.37}} & \hfil{\textbf{64.94}} \\
\bottomrule
\end{tabular}
\end{center}
\end{table}

The FG-NET dataset comprises 1,002 facial images of 82 subjects. Each subject has more than 10 facial images taken over a long time span. The MORPH II dataset is the largest among these three datasets and contains over 55,000 images of approximately 13,000 subjects. The age span in this dataset ranges from 16 to 77, with an average age of 33. The CACD-VS dataset includes 2,000 positive pairs and 2,000 negative pairs, where each positive pair consists of two images of the same subject, and each negative pair contains two images of different subjects. 

\begin{table}
\begin{center}
\label{tab:morph_homogenous}
\caption{Comparison between state-of-the-art methods and CACon for homogeneous-dataset evaluations on the MORPH II dataset.}
\begin{tabular}{l@{\hskip 0.5in}c@{\hskip 0.5in}c}\toprule
\hfil{Method} & \hfil{Setting-1} & \hfil{Setting-2} \\ \midrule
\hfil{LF-CNN \cite{wen2016latent}} & \hfil{97.51} & \hfil{-} \\
\hfil{OE-CNN \cite{wang2018orthogonal}} & \hfil{98.55} & \hfil{98.67} \\
\hfil{DM \cite{shakeel2019deep}} & \hfil{98.67} & \hfil{-} \\
\hfil{AIM \cite{zhao2019look}} & \hfil{99.13} & \hfil{98.81} \\
\hfil{DAL \cite{wang2019decorrelated}} & \hfil{98.93} & \hfil{98.97} \\
\hfil{MT-MIM \cite{hou2021disentangled}} & \hfil{-} & \hfil{99.43} \\
\midrule
\hfil{SimCLR \cite{chen2020simple}} & \hfil{94.08} & \hfil{93.79} \\
\hfil{CACon (ours)} & \hfil{\textbf{99.57}} & \hfil{\textbf{99.52}} \\
\bottomrule
\end{tabular}
\end{center}
\end{table}

\begin{table*}
\begin{center}
\label{tab:cross}
\caption{Comparison between AIFR methods and CACon for cross-dataset evaluations, where MO indicates the MORPH II dataset, FG indicates the FG-NET dataset, and CA indicates the CACD-VS dataset.}
\begin{tabular}{l@{\hskip 0.25in}c@{\hskip 0.25in}c@{\hskip 0.25in}c@{\hskip 0.25in}c@{\hskip 0.25in}c@{\hskip 0.25in}c}\toprule
\hfil{Method} & \hfil{MO $\Rightarrow$ FG} & \hfil{MO $\Rightarrow$ CA} & \hfil{FG $\Rightarrow$ MO} & \hfil{FG $\Rightarrow$ CA} & \hfil{CA $\Rightarrow$ MO} & \hfil{CA $\Rightarrow$ FG} \\ \midrule
\hfil{AIM \cite{zhao2019look}} & \hfil{77.68} & \hfil{81.87} & \hfil{53.39} & \hfil{60.59} & \hfil{71.11} & \hfil{51.61} \\
\hfil{MTLFace \cite{huang2021age}} & \hfil{79.64} & \hfil{81.93} & \hfil{58.18} & \hfil{60.63} & \hfil{73.00} & \hfil{52.03} \\
\midrule
\hfil{SimCLR \cite{chen2020simple}} & \hfil{76.04} & \hfil{79.44} & \hfil{54.07} & \hfil{58.40} & \hfil{63.94} & \hfil{46.76} \\
\hfil{CACon (ours)} & \hfil{\textbf{82.03}} & \hfil{\textbf{86.77}} & \hfil{\textbf{64.53}} & \hfil{\textbf{62.85}} & \hfil{\textbf{75.38}} & \hfil{\textbf{58.37}} \\
\bottomrule
\end{tabular}
\end{center}
\end{table*}

For fair comparisons, following previous works \cite{hou2021disentangled}, we utilize two large-scale facial datasets, MS-Celeb-1M \cite{guo2016ms} and CASIA-Webface \cite{yi2014learning}, to pre-train the identity feature extractor and also conduct cross-dataset experiments.

\subsection{Experiment Settings}



\textbf{Data Preparation:} Under the homogeneous-dataset settings, for the FG-NET dataset, we adopt three different settings for homogeneous-dataset evaluation. We first use the leave-one-image-out (LOIO) strategy as in previous works \cite{hou2021disentangled}. Specifically, 1 image is used for testing, and the remaining 1,001 images are used for fine-tuning. The whole process is repeated 1,002 times, and the average result is reported. We also adopt the protocols from Megaface challenge 1 (MF1) \cite{kemelmacher2016megaface} and Megaface challenge 2 (MF2) \cite{nech2017level}.

For the MORPH II dataset, we employ the partition strategy in \cite{zhao2019look, wang2018orthogonal, wang2019decorrelated}, where either 20,000 images from 10,000 subjects (Setting-1) or 6,000 images from 3,000 subjects (Setting-2) are used as the test set.





\textbf{Implementation Details:} We employ ResNet-50 as the function $f(\cdot)$ to extract comprehensive features from input images and a 3-layer fully-connected network to produce the features used by the modified NT-Xent loss. We use a batch size of 4,096 during pre-training, and 512 during fine-tuning. Additionally, we utilize the LARS optimizer for multi-GPU training during pre-training and Stochastic Gradient Descent (SGD) as the optimizer during fine-tuning.

\subsection{Comparison with State-of-the-Art Methods}
The comparison between recent AIFR methods and CACon under homogeneous-dataset settings on the FG-NET dataset is tabulated in Table 1. It is shown that CACon achieves comparable results compared to other state-of-the-art methods under the LOIO strategy. This is due to the fact that the face synthesis model we use is not designed to synthesize children faces and there is a large proportion of them in the FG-NET dataset. On the other hand, our model outperforms other methods by a large margin under settings MF1 and MF2. This is due to the fact that contrastive learning has the capability to extract more robust features for transfer learning and, hence, can achieve better performance on downstream tasks.

We also compare our method with SimCLR, which can be treated as the baseline model without the ability to learn age-invariant identity features during the pre-training stage. As can be seen from the table, by introducing the third samples within different age groups, CACon is able to achieve a higher accuracy by learning cross-age identity features. Moreover, since the synthesized sample is within a random age group, the model can disentangle both short-range and long-range age variations from the identity information.

Tables 2 shows the comparison between recent methods and CACon on the MORPH II dataset. Since the subjects in this datasets are either teenagers or adults, with the help of high-quality images from the face synthesis model, CACon can therefore achieve promising results.

We conduct cross-dataset experiments on all three datasets with one dataset as the source and one of the other as the target. Here, after pre-training, we fine-tune the model on the source dataset, and then evaluate on the target dataset directly without further fine-tuning. In this case, we only involve two open-source AIFR methods and the baseline model, i.e., SimCLR. The results are tabulated in Table 3. It is worth noting that due to the age span in the FG-NET dataset being much longer than that in the other two datasets, the accuracy, for example, from MORPH II to FG-NET is generally lower compared to that from MORPH II to CACD-VS. Although SimCLR does not learn age-invariant identity features explicitly, due to the better generalization ability gained from contrastive learning, it can achieve comparable results compared to recent AIFR methods. Further with modifications, our CACon outperforms state-of-the-art by a large margin.

\section{Conclusion}

In this paper, we proposed CACon to tackle the AIFR problem. Different from previous AIFR works, our method utilizes contrastive learning with an additional augmented sample generated by a face synthesis model to force the method to maximize the similarities among features from the facial images of the same subject within different age groups. Based on evaluations on several AIFR benchmark datasets, CACon achieves state-of-the-art performance in both homogeneous-dataset and cross-dataset evaluations.

\bibliographystyle{IEEEbib}
\bibliography{refs}

\end{document}